\documentclass{article}
\usepackage{amsmath}

\usepackage[preprint]{neurips_2025}

\usepackage[utf8]{inputenc} 
\usepackage[T1]{fontenc}    
\usepackage{hyperref}       
\usepackage{url}            
\usepackage{booktabs}       
\usepackage{multirow}       
\usepackage{amsfonts}       
\usepackage{nicefrac}       
\usepackage{microtype}      
\usepackage{xcolor}         
\usepackage{graphicx}
\setcitestyle{square,numbers}
\usepackage{caption}
\usepackage{authblk}
\usepackage{textcomp}
\usepackage{multirow}

\title{\textit{SafeMVDrive}: Multi-view Safety-Critical Driving Video Synthesis in the Real World Domain}
\author{
  Jiawei Zhou$^{1}$,
  Linye Lyu$^{1}$,
  Zhuotao Tian$^{1}$,
  Cheng Zhuo$^{2}$,
  Yu Li$^{2}$\thanks{Corresponding author: Yu Li at yu.li.sallylee@gmail.com}\\
  $^1$Harbin Institute of Technology, Shenzhen \\
  $^2$Zhejiang University
}
\begin{document}
\maketitle

%



\begin{abstract}
Safety-critical scenarios are rare yet pivotal for evaluating and enhancing the robustness of autonomous driving systems. While existing methods generate safety-critical driving trajectories, simulations, or single-view videos, they fall short of meeting the demands of advanced end-to-end autonomous systems (E2E AD), which require real-world, multi-view video data. To bridge this gap, we introduce SafeMVDrive, the first framework designed to generate high-quality, safety-critical, multi-view driving videos grounded in real-world domains.
SafeMVDrive strategically integrates a safety-critical trajectory generator with an advanced multi-view video generator. 
To tackle the challenges inherent in this integration, we first enhance scene understanding ability of the trajectory generator by incorporating visual context -- which is previously unavailable to such generator  -- and leveraging a GRPO-finetuned vision-language model to achieve more realistic and context-aware trajectory generation. 
Second, recognizing that existing multi-view video generators struggle to render realistic collision events, we introduce a two-stage, controllable trajectory generation mechanism that produces collision-evasion trajectories, ensuring both video quality and safety-critical fidelity. 
Finally, we employ a diffusion-based multi-view video generator to synthesize high-quality safety-critical driving videos from the generated trajectories.
Experiments conducted on an E2E AD planner demonstrate a significant increase in collision rate when tested with our generated data, validating the effectiveness of SafeMVDrive in stress-testing planning modules. Our code, examples, and datasets are publicly available at: \href{https://zhoujiawei3.github.io/SafeMVDrive/}{https://zhoujiawei3.github.io/SafeMVDrive/}.

\end{abstract}

\section{Introduction}
\label{introduction}


Vision-based end-to-end (E2E) autonomous driving (AD) systems, which directly map visual inputs to driving decisions, have gained growing attention and are gradually deployed in real-world environments \cite{zheng2024genad, hu2023_uniad,  li2024hydra, liao2024diffusiondrive, jiang2024senna}. However, ensuring their safety in diverse scenarios remains a significant challenge and addressing this issue requires large-scale and diverse driving datasets, especially the safety-critical ones. Yet, collecting such data in the real world is both costly and inherently dangerous. As a promising alternative, synthetic generation of safety-critical driving scenarios offers a scalable, low-risk, and cost-effective solution. These synthetic datasets can significantly enhance the robustness and generalization of E2E AD systems, especially in rare and hazardous conditions.

\begin{figure}[t]
 \vskip -0.2in
  \centering
  \centerline{\includegraphics[width=\columnwidth]{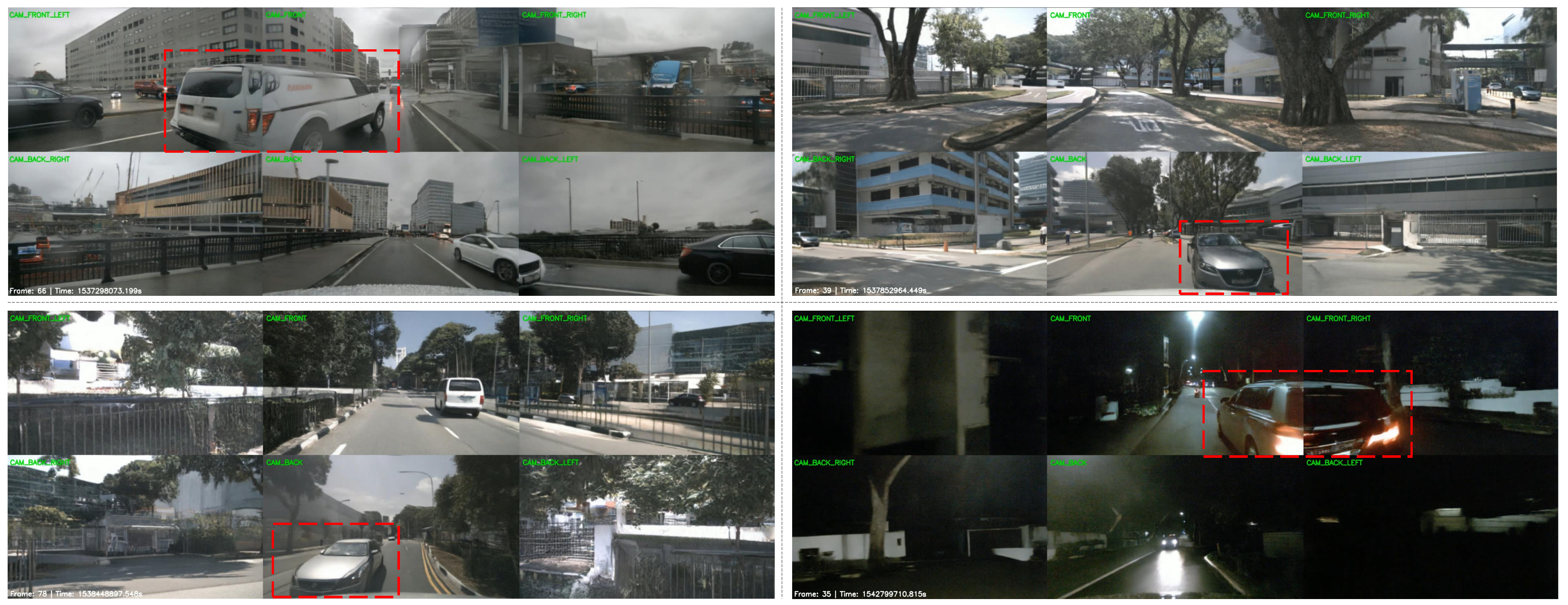}}
  \vskip -0.05in
  \caption{Keyframes from diverse realistic, multi-view, safety-critical videos generated by \textbf{SafeMVDrive}. Red boxes indicate safety-critical vehicles involved in events like cut-ins, rapid rear approaches, and sudden braking. Additional video examples are available via the link provided in the abstract.}
  \label{introduction}
  \vskip -0.2in
\end{figure}

Most existing methods for safety-critical data generation aim to generate realistic and controllable adversarial trajectories with diffusion models \cite{xu2025diffscene, zhong2023ctg, zhong2023CTG++,chang2024safesim}, which can be used to evaluate and improve the performance of the AD planning modules. However, these methods produce non-visual trajectories, which are incompatible with E2E AD systems that require visual input. While some approaches use simulators to generate safety-critical driving videos  \cite{zhang2024chatscene, abeysirigoonawardena2019generating}, their effectiveness is limited by the domain gap between simulation and reality. Besides, deep video generative models like Open-Sora \cite{opensora} are also explored to generate real-world driving accident videos,  but the produced videos are typically low-quality and limited to single-view outputs \cite{li2025AVD2}.

 With recent AD video models supporting realistic multi-view video generation and controllability via signals \cite{wen2024panacea, sima2024drivelm, chen2024unimlvg}, a naive method to generate multi-view safety-critical videos in the real domain is to convert the safety-critical trajectories—produced by the trajectory generator—into control signals to guide the video generator. However, this naive approach faces several challenges: firstly, current trajectory generators need to select an adversarial vehicle from multi-vehicle traffic, typically relying on heuristic rules based on non-visual data such as annotated vehicle kinematics and map features. Due to the inherent limitations of heuristic methods and lack of critical visual cues, it is difficult for the selector to comprehensively understand the complex physical scene to choose the appropriate vehicle (as detailed in Section \ref{3.2}). Consequently, the safety-criticality and realism of the generated videos can be compromised. 
 Furthermore, current safety-critical trajectory generators aim to generate collision events. However, existing multi-view video generators struggle to realistically simulate them due to a lack of multi-view collision training data. When the control signal corresponds to collision trajectories, the realism of the generated video degrades significantly.

To address the above issues, we present \textit{SafeMVDrive} to generate high-quality, multi-view safety-critical videos in the real-world domain. Our key insight lies in enhancing adversarial vehicle selection by incorporating visual information and simulating evasion trajectories that align with the capacity of existing multi-view video models. 
For adversarial vehicle selection, we leverage the strong scene comprehension capabilities of Vision-Language Models (VLMs) \cite{chen2024spatialvlm} and design a VLM-based selector to selects the most critical adversarial vehicle based on visual data from the initial scene. 
To adapt the VLM to the selection task, we first construct an automatically annotated dataset that maps scenes to set of vehicles capable of colliding with the ego vehicle. We then fine-tune the VLM using the GRPO algorithm \cite{shao2024deepseekmath}, enabling it to reason about complex traffic situations and significantly improve the success rate of adversarial vehicle selection. Furthermore, to address the limitations of video generators in rendering collisions, we introduce a two-stage trajectory generator. In the first stage, we simulate a valid collision trajectory. In the second stage, we refine this into a natural evasion trajectory that maintains safety-critical features while avoiding direct collisions. This ensures that the generated videos remain realistic and within the capability of current video models. Finally, by integrating a state-of-the-art multi-view video generator, we produce high-quality safety-critical driving videos in the real-world domain. As shown in Figure \ref{introduction}, our method successfully generates realistic traffic scenes suitable for testing and improving E2E AD systems.


The main contributions of our work are summarized as follows:
\begin{itemize}
    \item In this paper, we introduce \textbf{SafeMVDrive}, the first framework capable of generating high-quality, safety-critical multi-view video in the real-world domain. The core insight underlying our framework is the strategic integration of a safety-critical trajectory simulator with a multi-view driving video generator, and addressing the main challenge of their integration using a VLM-based adversarial vehicle selector and a two-stage evasion trajectory generator.
    \item We incorporate visual information into the selection of safety-critical vehicles by adapting a vision-language model (VLM) to this task. To facilitate this, we propose an automated annotation method to generate pairs of driving scenes and the corresponding safety-critical vehicles. Using this dataset, we fine-tune the VLM with the GRPO algorithm to improve its ability to understand multi-view driving scenes, allowing it to accurately identify adversarial vehicles capable of inducing safety-critical scenarios.
    \item We propose a two-stage trajectory generator to produce collision evasion trajectories that remain within the generative capacity of existing multi-view video generators. In the first stage, a collision trajectory is generated. In the second stage, we design a method to transform the first-stage collision trajectories to natural evasion trajectories, preserving the safety-critical characteristics while ensuring compatibility with video generation models.
    \item Using our framework, we construct the first high-quality, multi-view, safety-critical driving video dataset in the real domain. The dataset contains 41 diverse 9-second scenes and can serves as a valuable benchmark to evaluate and enhance the robustness of end-to-end autonomous driving planners in terms of their ability to avoid collisions.

\end{itemize}

Our dataset exhibits strong safety-critical traits. Tested with the classic end-to-end autonomous planner UniAD \cite{hu2023_uniad}, our generated videos show a 30\% increase in collision rate compared to the original NuScenes videos \cite{caesar2020nuscenes}. Moreover, our videos maintain high visual realism. In the user study evaluating video quality, our safety-critical videos achieve 87\% of the realism score compared to those generated by the state-of-the-art video generator using original, non-safety-critical trajectories.


\section{Related work}
\label{related_work}

Safety-critical data generation is essential to enhance end-to-end AD systems' robustness in the real world. The existing work can be categorized into trajectory-based and video-based approaches in terms of their output formats. Trajectory-based approaches generate non-visual adversarial trajectories, while video-based approaches produce safety-critical driving videos.

Given the initial traffic context, trajectory approaches first select the adversarial vehicle and then optimize trajectories that lead to safety-critical situations. Recent works often leverage diffusion models for controllable traffic generation. For example, Controllable Traffic Generation (CTG) trains diffusion models on large-scale driving data to generate realistic trajectories \cite{zhong2023ctg}. Safe-Sim \cite{chang2024safesim} and CTG++ \cite{zhong2023CTG++} further select adversarial vehicles via heuristics and apply adversarial losses to guide generation. While these methods show progress, they are incompatible with E2E AD systems, which require visual input. Besides, they use simple heuristic methods like the nearest vehicle to select the adversarial vehicle, which can fail to create safety-critical cases. For instance, as shown in Figure \ref{introduceVLM}, a nearby bus may be chosen despite being blocked by obstacles, making collision impossible.

Another research direction is to directly provide safety-critical visual data. Some works \cite{zhang2024chatscene, abeysirigoonawardena2019generating, xu2025diffscene} use simulators like Carla \cite{Dosovitskiy2017carla} to generate adversarial driving videos but suffer from the domain gap between simulation and reality. To generate realistic visuals, ADV2 \cite{li2025AVD2} employs generative models like open-sora \cite{opensora}, finetuned on real traffic accident data with text captions, to produce adversarial videos from user prompts. However, the videos are low-quality and single-view, limiting their use in E2E AD systems that require high-quality, multi-view inputs.

In contrast to the current works, our proposed framework strategically combines a novel vehicle selector, an evasion trajectory generator, and a high-performance driving video generator to produce realistic, high-quality, multi-view adversarial driving video data compatible with E2E AD systems.

\section{Methods}
\label{methods}


\begin{figure}[t]
  \vskip -0.1in
  \centering
  \centerline{\includegraphics[width=\columnwidth]{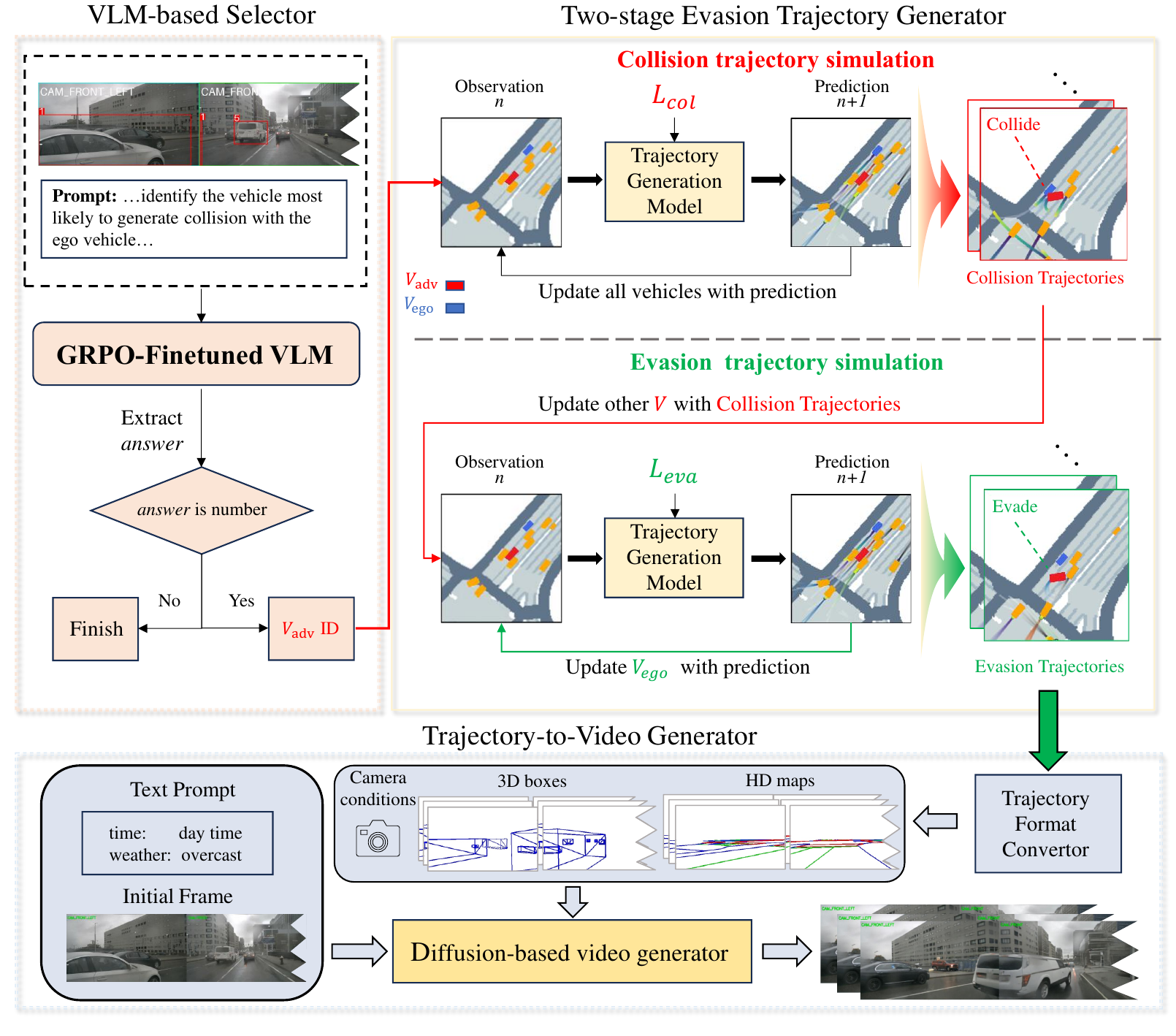}}
  \vskip -0.1in
  \caption{The SafeMVDrive framework for generating realism, multi-view safety-critical videos. }
  \label{pipeline}
  \vskip -0.2in
\end{figure}

\subsection{Overview}
Figure \ref{pipeline} shows our framework for generating safety-critical multi-view videos, comprising three parts: (1) a VLM-based adversarial vehicle selector; (2) a two-stage evasion trajectory generator; and (3) a trajectory-to-video generator. The input is single-frame holistic information of an initial scene, combining visual data (multi-view camera images) and non-visual data (camera parameters, vehicle states, and road maps)—all available in datasets like NuScenes \cite{caesar2020nuscenes}, Waymo \cite{sun2020waymo}, and Argoverse2 \cite{wilson2023argoverse}. First, we mark vehicles within distance \(D\) from the ego vehicle with ID-labeled 2D boxes in the multi-view images. The images are then fed into the VLM-based selector to identify the adversarial vehicle \(V_{adv}\). With  \(V_{adv}\)'s ID, the two-stage evasion trajectory generator can produce safety-critical trajectories. In the first stage, we generate a collision trajectory where \(V_{adv}\) collides with the ego vehicle \(V_{ego}\); in the second stage, we convert this collision trajectory into a realistic evasion trajectory using our proposed method. The generated trajectories are then converted to control signals that guide a diffusion-based video generator to synthesize realistic safety-critical multi-view videos.



\subsection{VLM-based Adversarial Vehicle Selector}\label{3.2}

\begin{figure}[t]

  \centering
  \centerline{\includegraphics[width=\columnwidth]{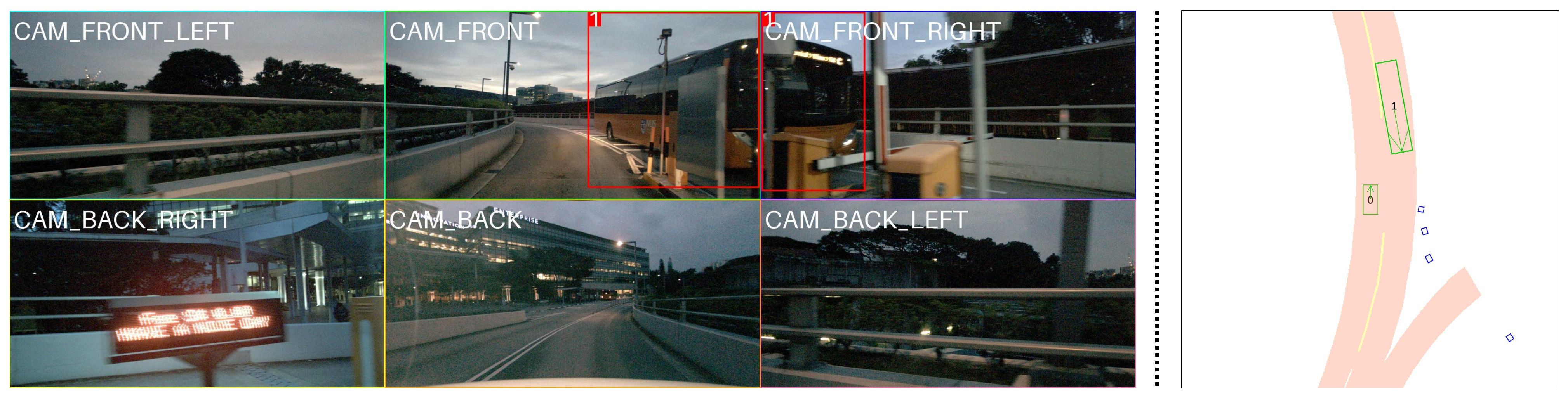}}
  \vskip -0.1in
  \caption{Comparison between the real-world scene (left) and the BEV-rendered non-visual data (right). Obstacles that physically prevent a collision between Vehicle 1 and the ego vehicle are visible in the real-world view but missing in the non-visual data, potentially misleading heuristic methods.}
  \label{introduceVLM}
  \vskip -0.2in
\end{figure}

The first step in generating safety-critical data is selecting the adversarial vehicle \(V_{adv}\)
  from the initial scene. Prior methods rely on simple heuristics using non-visual data like vehicle kinematics and maps, such as choosing the closest vehicle, applying fixed distance-velocity rules \cite{zhong2023CTG++}, or randomly picking nearby lane vehicles \cite{chang2024safesim}. However, these heuristics lack crucial visual cues and fail to capture complex driving scenarios, often resulting in inappropriate selections. Figure \ref{introduceVLM} illustrates this: from the BEV-view, non-visual data misses an obstacle separating the chosen vehicle from the ego vehicle, causing all heuristic methods to select 
\(V_{adv}\) incorrectly (red boxes).



To address the aforementioned problems, we propose incorporating visual information into adversarial vehicle selection by leveraging the scene understanding capabilities of Vision-Language Models (VLMs) \cite{chen2024spatialvlm}. Specifically, we introduce a VLM-based selector that selects the critical adversarial vehicle using the multi-view images from the initial scene.
Our first attempt is to guide the VLM with task-specific prompts. To aid comprehension and accurate vehicle ID output, we annotate safety-critical vehicle candidates with ID-labeled 2D bounding boxes (Figure \ref{introduceVLM}, left) and exclude distant vehicles (beyond distance 
\(D\) from the ego vehicle). We also design a tailored prompt using these annotations (see Appendix \ref{label detailed}). However, due to VLMs’ limited exposure to multi-view data during training, prompting alone proves insufficient for effective multi-view understanding.

To address the above problem, we fine-tune the VLM to better adapt it to our task, which requires constructing a suitable fine-tuning dataset. A key challenge is determining the correct VLM output for each multi-view image—specifically, identifying which vehicles could realistically collide with the ego vehicle via natural trajectories. Manual labeling is costly and error-prone. To solve this, we propose an automated method using a controllable diffusion-based traffic simulator \cite{zhong2023ctg} that generates naturalistic trajectories and supports test-time guidance via loss functions. For each safety-critical vehicle candidate \(V_{adv}\), we apply a loss encouraging collision with the ego vehicle based on their distance (see Section \ref{3.3}). We select vehicles that successfully collide and filter out unrealistic cases, such as those entering non-drivable areas or colliding with other vehicles first. This yields annotated data defining the set of effective safety-critical vehicles \(S_{coll}\) for each scene.

After obtaining the fine-tuning dataset, we apply the GRPO algorithm \cite{shao2024deepseekmath}, a recent RL method proven to effectively enhance reasoning of LLMs \cite{guo2025deepseek} and VLMs \cite{shen2025vlm}. GRPO enhances the model's reasoning capabilities through a self-improving RL process, which makes it well-suited for helping the model better understand complex multi-view physical scenarios \cite{wei2022chain,ho2022large}. Following \cite{shen2025vlm}, we augment the prompt with: `Output the thinking process in <think> </think> and final answer (number) in <answer> </answer> tags.' and use the following format reward:
\begin{equation}
R_{form}= 
\begin{cases} 
1 & \text{if } O \sim ⟨ \texttt{think} ⟩ ... ⟨ /\texttt{think} ⟩ \; ⟨\texttt{answer} ⟩ ... ⟨ / \texttt{answer} ⟩ \\
0 & \quad \quad \quad  \quad \quad \quad \quad  \  \quad  \quad \; \quad  \quad \quad \quad \quad \quad \quad \quad \quad \quad  
 \text{otherwise}
\end{cases}
\end{equation}
where \(O\) refers to the output from the VLM. The format reward is designed to enforce the model to place its reasoning process between `<\texttt{think}>’, and outputs the final answer within the `<\texttt{answer}>' tags. To promote accurate outputs, we add the following accuracy reward,
\begin{equation}
R_{Acc} =
\begin{cases}
similarity(extract\_answer(O), \text{"no vehicle is appropriate"}) \quad\text{if } S_{\text{coll}} = \emptyset \\
1 \quad \quad \quad \quad \  \quad \quad \quad   \quad  \; \; \quad \quad  \quad \quad    \text{if } S_{ \text{coll}} \neq \emptyset \land extract\_answer(O) \in S_{\text{coll}} \\
0 \text{\quad \quad \quad  \quad \quad \quad \quad \quad \quad \quad \quad \quad \quad  \quad \quad \quad \quad \quad \quad \quad \quad \quad \quad \quad \quad \quad } \text{otherwise}
\end{cases}
\end{equation}

where \(extracted\_answer()\) refers to extracting the content between the tags <\texttt{answer}>...</\texttt{answer}> from the VLM output. \( S_{\text{coll}} \) denotes the set of vehicles that can collide with the ego vehicle in our automated annotation process. If \( S_{\text{coll}} = \emptyset \), it indicates that no vehicle in the scene can collide with the ego vehicle, in which case we expect the model to output "no vehicle is appropriate". If \( S_{\text{coll}} \ne \emptyset \), at least one adversarial vehicle exists, and the model should output an ID belonging to this set.


\subsection{Two-stage Evasion Trajectory Generator}
\label{3.3}

Several safety-critical trajectory simulators have been introduced \cite{xu2025diffscene, zhong2023CTG++, chang2024safesim}, primarily focusing on generating collision events. However, current multi-view video generators struggle to realistically generate such events, degrading visual quality when control signals cause collisions. To address this, we propose a two-stage evasion trajectory generator. It produces safety-critical yet non-colliding evasion trajectories compatible with current video generators while retaining safety-critical features.

Our generator builds upon a popular controllable trajectory generation framework \cite{zhong2023ctg}, which uses a diffusion-based model trained on real-world driving data for realistic trajectories. It enables test-time control via loss functions.  Since our framework takes a single-frame initial scene as input, we retrain the trajectory generation model to align with this setup. We adopt the closed-loop simulation strategy described in \cite{zhong2023ctg}. At each step \(n\), the model predicts future trajectories from the current scene, applying only the first few to update the scene. This iterates to form the full trajectory sequence.

Once the VLM-based selector identifies the adversarial vehicle \(V_{adv}\), our trajectory generator performs a two-stage simulation. In the first stage,  \(V_{adv}\) is guided to collide with \(V_{ego}\). If the collision occurs before  \(V_{adv}\) entering non-drivable areas or hitting others, the collision trajectory is considered valid. In the second stage, we introduce a trajectory update mechanism with an evasion-targeted loss, which guides \(V_{ego}\) in evading \(V_{adv}\). Finally, a collision-evasion trajectory sequence is generated.

During the collision-stage trajectory simulation, we employ three loss functions for test-time guidance: an adversarial loss, a no-collision loss, and an on-road loss. The adversarial loss is necessary to encourage \(V_{adv}\) to collide with \(V_{ego}\), typically based on their distance \cite{zhong2023CTG++, chang2024safesim}. However, this often causes \(V_{adv}\) to remain stuck to \(V_{ego}\) after collision, resulting in unnatural dynamics—shifting of \(V_{adv}\)  from aggressive (e.g., rapid acceleration) to passive, ego-like behaviors (e.g., slow driving). To solve this, we propose the following adversarial loss formulation:
\begin{equation}
L_{adv} = 
\begin{cases} 
\displaystyle \sum_{t=1}^{T} w_t \cdot d_t \cdot \mathbb{I}(d_t > d_{penalty}) & \text{Before } V_{adv} \text{ collides with } V_{ego} \\
0 & \text{After } V_{adv} \text{ collides with } V_{ego}
\end{cases}
\end{equation}
where \(T\) denotes predicted future steps, \(d_t\) is the distance between \(V_{adv}\) and \(V_{ego}\) at time step \(t\), and \(d_{\mathrm{penalty}}\) is the non‐collision distance threshold. Gradients are detached with respect to \(V_{ego}\) to ensure only \(V_{adv}\) has the adversarial behavior. A time-decay weight  \(w_t = \frac{\lambda^t}{\sum_{k=0}^{T-1} \lambda^k}\), controlled by a decay factor \(\lambda\), emphasizes earlier trajectory predictions and is shared across all losses. Moreover, to avoid unnatural sticking post-collision, we explicitly set the adversarial loss \(L_{adv}\) to zero once the collision between \(V_{adv}\) and \(V_{ego}\) has occurred in the updated trajectories during closed-loop simulation.  This leads to more natural post-collision behavior of the adversarial vehicle.

To prevent undesired collisions (excluding that between the ego and adversarial vehicles), we utilize a no-collision loss \(L_{no\_coll}\), which penalizes inter-vehicle collisions in denoised trajectories, excluding the ego–adversarial pair. To keep vehicles on drivable areas, an on-road loss \(L_{on\_road}\) penalizes trajectories entering non-drivable zones and guides them back. Full definitions of \(L_{no\_coll}\) and \(L_{on\_road}\) can be found in Appendix \ref{loss}.

Overall, the loss function of the collision stage trajectory simulation can be summarized as
\begin{equation}
L_{coll} =  \alpha L_{adv} + \beta L_{no\_coll} + \gamma L_{on\_road}
\end{equation}
where $\alpha$\label{alpha}, $\beta$\label{beta}, $\gamma$\label{gamma} control each loss’s contribution. We obtain trajectory sequences through closed-loop simulation and we filter out trajectories that either do not collide with the ego vehicle or that collide with other vehicles or go off-road beforehand to ensure safety-criticality and physically validity. Subsequently, the collision trajectories are fed into the second stage for evasion trajectory simulation.

During the evasion stage trajectory simulation, we only use \(L_{\text{no\_coll}}\) and \(L_{\text{on\_road}}\) as follow, 
\begin{equation}
L_{eva} = \beta L_{no\_coll} + \gamma L_{on\_road}
\end{equation}
where the \(L_{\text{no\_coll}}\) is applied to all vehicles in the scene to guide \(V_{ego}\)  in evading \(V_{adv}\). The evasion-stage simulation starts from the same initial scene as the collision stage. During closed-loop rollout, only the ego vehicle’s trajectory is updated via the diffusion model; other vehicles retain their collision-stage trajectories to preserve adversarial behavior. This converts a collision scenario to a safety-critical evasion one, staying within the video generator’s capability. Finally, we select successful evasive trajectories for video generation.

\begin{table}[t]
\caption{Comparison of baseline effectiveness by evaluating video realism and planner's collision rate ($CR$) on the videos. Sample-level ($CR$) measures the average collision probability per valid sample, while scene-level ($CR$) counts the average number of colliding valid samples per scene.}
\label{table1}
\begin{center}
\begin{small}
\resizebox{\columnwidth}{!}{
\begin{tabular}{lcccccccccc}
\toprule
\multirow{2}*{\textbf{Methods}} & \multicolumn{4}{c}{Sample-level $CR$ $\uparrow$} & \multicolumn{4}{c}{Scene-level $CR$ $\uparrow$} &\multirow{2}*{FID $\downarrow$}& \multirow{2}*{Natural score $\uparrow$} \\ 
\cmidrule(lr){2-5}\cmidrule(lr){6-9}
                         & 1s            & 2s     & 3s & Avg.   & 1s            & 2s     & 3s & Avg.   &     &          \\ 
\midrule
Origin & 0.001 & 0.003 & 0.007 & 0.004 & 0.024 & 0.049 & 0.122 & 0.065 & 16.254 & $0.633\pm0.133$ \\
Naive & 0.082 & 0.274 & 0.445 & 0.267 & 0.556 & 1.861 & 3.028 & 1.815 & 23.346 & $0.207\pm0.183$ \\
SafeMVDrive & 0.097 & 0.207 & 0.303 & 0.202 & 1.659 & 3.561 & 5.220 & 3.378 & 20.626 & $0.560\pm0.122$ \\
\bottomrule
\end{tabular}
}
\end{small}
\end{center}
\end{table}

\begin{figure}[t]
  \centering
  \centerline{\includegraphics[width=\columnwidth]{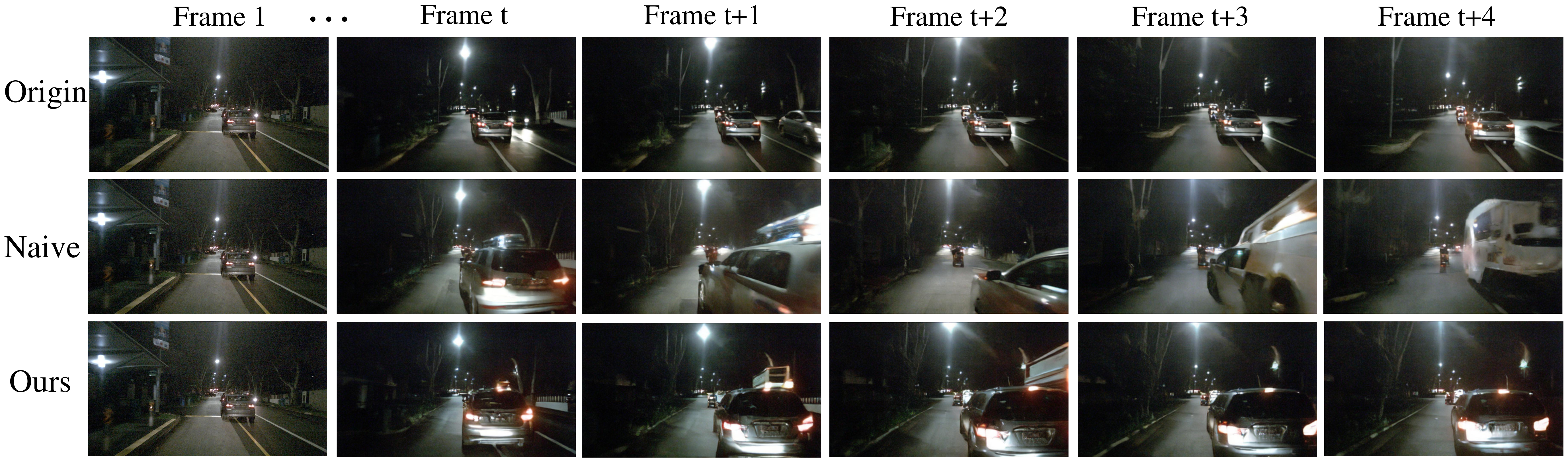}}
  \vskip -0.1in
  \caption{Comparison of videos generated by different methods, only showing front view. Origin is ordinary, Naive loses realism near the end, while only ours exhibits both realism and safety-criticality. }
  \label{compare}
  \vskip -0.2in
\end{figure}

\subsection{Trajectory-to-Video Generator}
\label{3.4}
To convert the simulated collision evasion trajectories into multi-view driving videos, we leverage diffusion-based video generation models tailored for autonomous driving scenarios. To ensure that the generated videos accurately reflect our trajectory-based traffic scenarios, we require a video generation model that explicitly encodes both the ego vehicle’s motion and the surrounding vehicles’ motion. Moreover, we require a model capable of producing sufficiently long video sequences as safety-critical scenarios
 typically spans a relatively long duration. Accordingly, we choose UniMLVG \cite{chen2024unimlvg} as our backbone. UniMLVG delivers state-of-the-art video quality and supports motion control signals like 3D bounding boxes, HD maps, and camera conditions. Its multi-task training scheme also reduces autoregressive errors, enabling high-quality long-video synthesis.


We first convert generated trajectories into frame-level control inputs—3D boxes, HD maps, and camera conditions—combined with multi-view initial frames and time-weather text to guide the video generation. Due to the extended duration of the collision evasion scenarios, we use an autoregressive roll-out to generate videos: the final frame of each roll-out serves as the initial frame for the next, and the corresponding control signals for the new time window are used to guide the subsequent generation. Through this iterative process, the complete collision evasion trajectories are ultimately transformed to a multi-view video in the real domain.

\section{Experiments}

\subsection{Experimental Settings}\label{4.1}

\textbf{Datasets:} We use the large-scale real-world NuScenes dataset \cite{caesar2020nuscenes}, featuring diverse driving scenarios.  To train the VLM for our adversarial vehicle selector, we randomly select 1,500 samples from the training split and generate the safety-critical annotations within each scene with the automated annotation method (proposed in Section \ref{3.2}). The trajectory generation diffusion model is trained on the full training split. For evaluation, 250 samples are randomly selected from the validation split.

\textbf{Baseline:} 
As the first to generate multi-view, realistic, safety-critical videos, we design two intuitive baselines for comparison. \textbf{Naive}, generates safety-critical videos by converting collision trajectories into control signals for the video generator. These trajectories are produced using the vehicle selection method and loss function from \cite{zhong2023CTG++}, combined with our retrained conditional diffusion-based model. \textbf{Origin}, uses original NuScenes trajectories to benchmark video quality under natural conditions. All baselines generate videos via UniMLVG with identical settings, as detailed in Section~\ref{3.4}.

\textbf{Metrics:} To demonstrate that the videos generated by our framework present significant challenges to end-to-end planners and are likely to induce collisions, we evaluate the classical end-to-end planner, UniAD \cite{hu2023_uniad} on our generated data. Following the methodology proposed by Li et al. \citep{li2024ego}, we compute the collision rate for each video sample. Specifically, the collision rate is defined as:

\begin{equation}
cr(t) = \left( \sum_{i=0}^{N} \mathbb{I}_i \right) > 0,\quad N = \frac{t}{0.5}.
\end{equation}
where $N$ denotes the number of the trajectory points in the planning before $t$ seconds, and $\mathbb{I}_i$ indicates whether the ego vehicle collides at step $i$. The final collision rate \(CR(t)\) is averaged over all samples. We follow this method with one adaptation: if a collision already occurs at the initial step ($\mathbb{I}_0=1$), the collision ratio cr(t) for this sample is always 1, making it unsuitable for evaluating planner performance. Therefore, we consider such samples as invalid initializations and exclude them from our evaluation. We report both sample-level and scene-level averaged \(CR(t)\): the former represents the averaged \(cr(t)\) before time t for each valid sample; the latter reflects the average number of valid samples in which the planner collides before time \(t\) within each scene.


\begin{figure}[t]
  \centering
  \centerline{\includegraphics[width=\columnwidth]{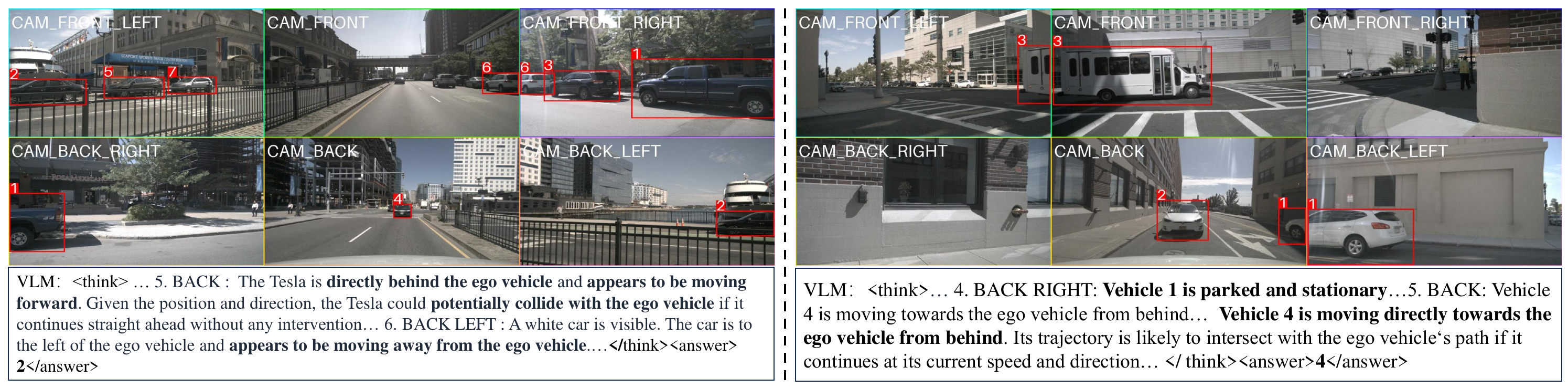}}
  \vskip -0.05in
  \caption{Adversarial vehicle selection examples using the GRPO-finetuned VLM. The VLM accurately analyzes spatial relationships between vehicles and makes reasonable selections.}
  \label{vlm_example}
  \vskip -0.2in
\end{figure}

Regarding the realism of generated videos, existing automatic metrics such as FVD \cite{unterthiner2018towards} are widely recognized as insufficient for accurately reflecting perceptual quality and real-world dynamics \cite{gao2024vista,bar2024lumiere, blattmann2023align, girdhar2024factorizing, wu2024towards}. Consequently, we rely on human evaluation to obtain a more reliable and authentic assessment. In line with recent studies \cite{gao2024vista,bar2024lumiere, blattmann2023align, blattmann2023stable,chen2023videocrafter1,wang2025lavie}, we employ the Two-Alternative Forced Choice (2AFC) protocol to evaluate the videos (see Appendix \ref{Human Evaluation} for details) and refer to the resulting preference rate as the realism score in our experiments. In addition, we also compute FID as an auxiliary quantitative metric to facilitate comparison with prior works.

\textbf{Implement Details}: We choose Qwen2.5VL-7B-Instruct~\cite{qwen2.5-VL} as our base VLM due to its strong vision-language understanding capabilities (fine-tuning details in Appendix~\ref{VLM-finetuning}). For safety-critical candidate selection, we set the distance threshold \(D = 25\,\text{m}\). We retrain the trajectory generation model to align with our single-frame input setup (details in Appendix \ref{VLM-finetuning}). In the two-stage simulation process, we set \(\alpha = 1\), \(\beta = 50\), \(\gamma = 1\) in the collision stage, \(\beta = 1\), \(\gamma = 1\) in the evasion stage, and use \(\lambda = 0.9\) for all loss terms. Ablation studies on these hyperparameters can be found in Appendix \ref{loss ablation study} and \ref{hyperparemeters of losses}. Our video generator produces 19 frames per iteration, using the last frame of the previous roll-out as the reference frame for the next. This results in a final video of 9 seconds at 12\,Hz.

\subsection{Evaluation of SafeMVDrive}\label{4.2}

This section compares the realism of videos generated by different baselines and the collision rate (CR) of the UniAD planner on these videos. Each method generates videos from 250 samples randomly selected from the NuScenes validation split (see Appendix \ref{dataset} for details). Table \ref{table1} shows SafeMVDrive videos significantly increase the planner’s CR while maintaining realism comparable to Origin. Specifically, average sample-level CR rises by nearly 0.2, and scene-level CR by 3.3, challenging the planner due to aggressive adversarial trajectories and ego evasion causing speed and acceleration variations. Origin achieves high realism but lacks safety-critical events, thus failing to challenge the planner. Naive’s scene-level CR is much lower (1.8 vs. 3.4) and its realism is only about half of SafeMVDrive’s, as collision events exceed the video generator’s capacity. Its slightly higher sample-level CR stems from vehicles getting stuck post-collision, increasing invalid samples and average CR. All methods show similar FID scores, indicating comparable image quality. Figure \ref{compare} compares the videos, highlighting SafeMVDrive as the only method combining high realism with strong safety-critical features.

\subsection{Evaluation of Our VLM-based Adversarial Vehicle Selector}

In this section, we evaluate the effectiveness of our VLM-based adversarial vehicle selector. On 250 validation scenes, we use automated annotation to identify all vehicles that can collide with the ego vehicle. We compare precision, recall, and F1-score of our VLM-based selector against three heuristic methods: Closest Vehicle, Rule-based Selector \cite{zhong2023CTG++}, and Random Adjacent \cite{chang2024safesim}. As shown in Table~\ref{table2}, our method achieves the highest F1-score, demonstrating its effectiveness in accurately identifying safety-critical vehicles. Figure \ref{vlm_example} shows examples of adversarial vehicle selections, where our VLM correctly analyzes positional relationships and driving directions to make appropriate selections.

We further evaluate the effectiveness of our GRPO fine-tuning. As shown in Table~\ref{table3}, the GRPO-finetuned model significantly outperforms the untuned baseline, achieving an F1-score improvement of 0.21 over the strongest 72B base model. We also evaluate supervised fine-tuning (SFT) for comparison (see Appendix~\ref{VLM-finetuning} for configurations), but it performs worse than GRPO, with an F1-score reduction of more than 0.05. These findings highlight both the necessity and effectiveness of adopting GRPO for our adversarial vehicle selection task.




\begin{table}[t]
  \centering
  \begin{minipage}{0.48\columnwidth}
    \centering
    \captionof{table}{Comparison of different methods for adversarial vehicle selection.}
    \label{table2}
    \resizebox{\linewidth}{!}{
    \begin{tabular}{lccc}
      \toprule
      \textbf{Methods}         & Precision         & Recall            & F1-score  \\ 
      \midrule
      Closest vehicle                   & 0.528              & 0.861       &  0.654 \\ 
      Rule-based selector            & 0.758      & 0.497 &0.600\\ 
      Random Adjacent      & 0.606   &  0.437 &0.507\\  
      VLM-based selector & 0.750 & 0.675 &\textbf{0.710} \\
      \bottomrule
    \end{tabular}
    }
  \end{minipage}
  \hfill
  \begin{minipage}{0.48\columnwidth}
    \centering
    \captionof{table}{Comparison of the performance of different models on adversarial vehicle selection.}
    \label{table3}
    \resizebox{\linewidth}{!}{
    \begin{tabular}{lccc}
      \toprule
      \textbf{Model}         & Precision         & Recall         & F1-score  \\ 
      \midrule
      Base 3B                  & 0.360              & 0.596      & 0.449           \\ 
      Base 7B            & 0.455       & 0.530& 0.489\\ 
      Base 72B       & 0.433   &  0.602 & 0.504\\  
      SFT-finetuned Model 7B  &0.582  &0.748 &0.655 \\ 
      GRPO-finetuned Model 7B & 0.750 & 0.675 &\textbf{0.710} \\
      \bottomrule
    \end{tabular}
    }
  \end{minipage}
\end{table}



\begin{table}[t]
\vskip -0.2in
\caption{Evaluation of the effectiveness of the two-stage simulation.}
\vskip -0.05in
\label{table4}
\begin{center}
\begin{small}
\begin{sc}
\resizebox{\columnwidth}{!}{
\begin{tabular}{lcccccccccc}
\toprule
\multirow{2}{*}{\textbf{Methods}} & \multicolumn{4}{c}{Sample-level $CR$ $\uparrow$} & \multicolumn{4}{c}{Scene-level $CR$ $\uparrow$} & \multirow{2}{*}{FID $\downarrow$} & \multirow{2}{*}{Natural score $\uparrow$}  \\ 
\cmidrule(lr){2-5}\cmidrule(lr){6-9}
                         & 1            & 2     & 3   & Avg.     & 1            & 2     & 3    & Avg.  &    &        \\ 
\midrule
Origin                   & 0.001              & 0.003         & 0.007 & 0.004  & 0.024              & 0.049         & 0.122 & 0.065 & 16.254 & 0.601 $\pm$ 0.134                  \\ 
Collision stage only    & 0.058         & 0.167       & 0.228 & 0.151  & 0.707            & 2.049     & 2.805 & 1.854 & 22.204 & 0.330 $\pm$ 0.128            \\ 
Two-stage simulation  & 0.097 & 0.207     & 0.303 & 0.202  & 1.659          & 3.561     & 5.220 & 3.378     & 20.626   & 0.569 $\pm$ 0.075 \\ 
\bottomrule
\end{tabular}
}
\end{sc}
\end{small}
\end{center}
\vskip -0.3in
\end{table}

\subsection{Evaluation of the Effectiveness of the Two-stage Simulation}\label{4.4}
We propose a two-stage trajectory simulator to generate collision-evasion scenarios that are both safety-critical and within the capability of current multi-view video generators. In this section, we assess the necessity of the two-stage simulation by comparing videos from collision-stage-only trajectories with those from the full two-stage process. Additionally, to assess the naturalness of the generated videos, we include the Origin baseline for comparison. Each method generates videos based on 250 samples randomly selected from the NuScenes validation split. Detailed generation procedures can be found in Appendix \ref{dataset}. As shown in Table \ref{table4}, our two-stage simulation leads to both higher collision rates for the planner and significantly improved realism.


\section{Conclusion}

We present \textit{SafeMVDrive}, the first framework for generating multi-view safety-critical driving videos in the real-world domain. By strategically combining a safety-critical trajectory simulator with a realistic multi-view video generator, we build a bridge from safety-critical trajectory simulation to multi-view video generation. To address the integration challenge, we introduce a VLM-based adversarial vehicle selector and a two-stage collision-evasion trajectory generation strategy. Experiments demonstrate the effectiveness of our approach in producing realistic and safety-critical multi-view videos, which lead to a high collision rate for end-to-end planners. The generated video data can serve as valuable resources for evaluating and enhancing autonomous driving systems.

\textbf{Limitations.}\label{5.1} Since this is the first work to generate multi-view safety-critical driving videos in the real world, we have several limitations. One is the reliance on the complete initial scene configuration, which restricts its ability to generate scenarios directly from raw multi-view camera inputs. Additionally, although our framework uses guidance signals to generate annotations, it lacks a mechanism to discard outdated or irrelevant ones—e.g., vehicles that have exited the ego’s view. Future research could address these challenges by reducing dependency on dense annotations and incorporating dynamic filtering strategies to maintain temporal relevance in the guidance signals.


\textbf{Societal Impact.}\label{5.2} This work aims to improve the safety and robustness of autonomous driving systems by generating realistic, safety-critical driving scenarios for testing and training. The potential for misuse is limited, as the primary application—autonomous driving—rarely involves malicious intent.

\bibliographystyle{acl_natbib}
\bibliography{neurips_2025}

\newpage
\appendix

\section{User Study Setting}\label{Human Evaluation}
\begin{figure}[t]
  \centering
  \centerline{\includegraphics[width=\columnwidth]{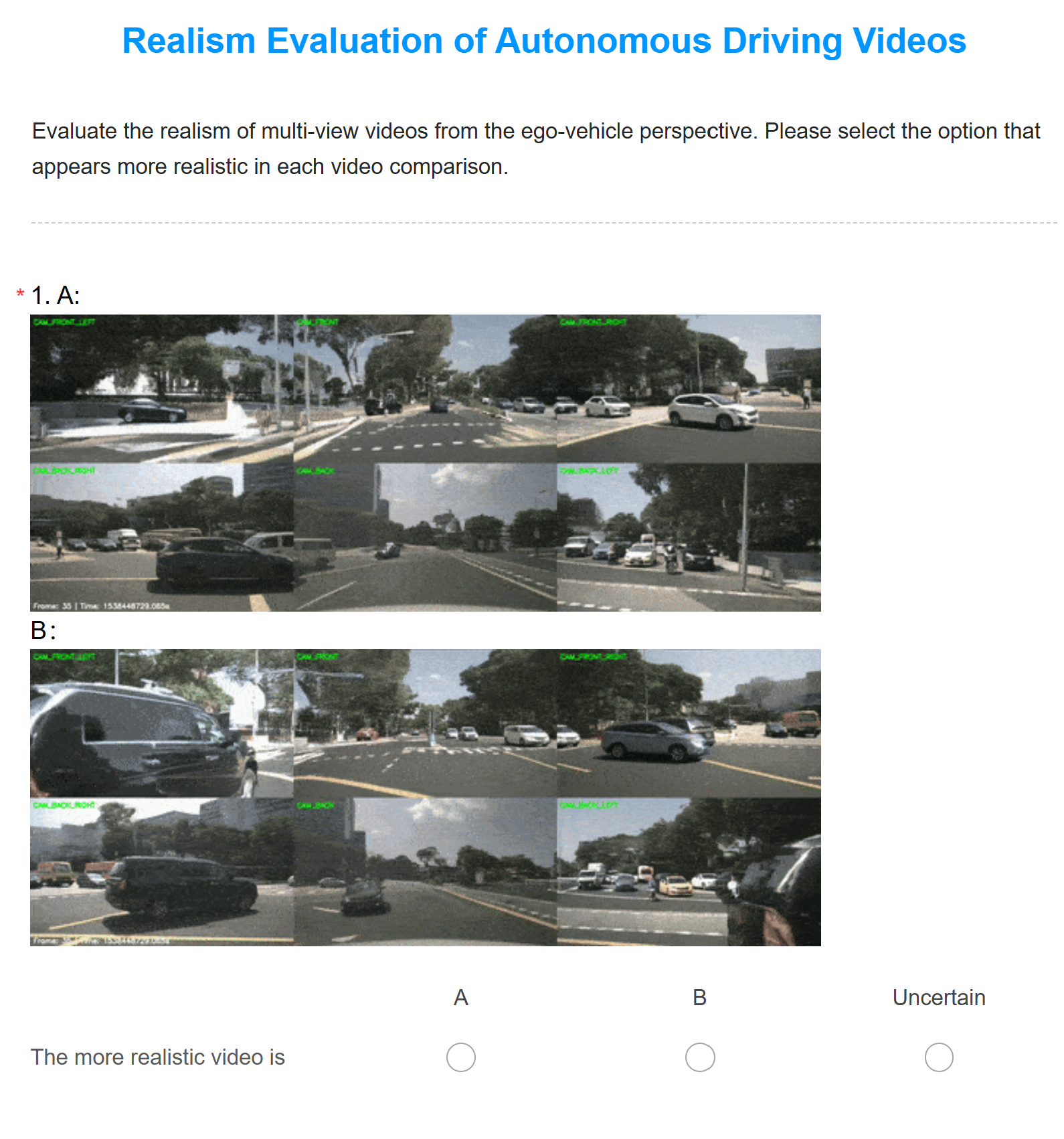}}
  \caption{The questionnaire used to evaluate the realism of videos generated by different baselines in the user study.}
  \label{human}
\end{figure}

In our experiments, participants are presented with two videos displayed side-by-side and are asked to choose the one they perceive to be of higher visual quality. In addition to choosing one of the two videos, an 'uncertain' option is also provided. A selected video receives 1 point; in the case of an "uncertain" response, both videos receive 0.5 point each. The final realism score is computed as the total number of points received divided by the total number of comparisons. The questionnaire we used is shown in Figure \ref{human}. The user studies in Section \ref{4.2} and Section \ref{4.4} are conducted separately. In the experiment of Section \ref{4.2}, we randomly select ten initial scenes that are present across all three video sets—Origin, Naive, and SafeMVDrive. For each selected scene, we retrieve the corresponding video from each set, forming ten matched triplets for pairwise comparison. Similarly, in Section \ref{4.4}, we randomly select ten initial scenes that exist in all three sets—Origin, Collision Stage Only, and Two-Stage Simulation—and obtain the corresponding video per method for each scene, again resulting in ten matched triplets for evaluation. For each user study, we collect 660 answers from 22 participants.

\section{No-collision Loss and On-road Loss}\label{loss}
To prevent collisions between the vehicles in the scene (except between the ego vehicle and adversarial vehicles), we use the following no-collision loss,
\begin{equation}
L_{no\_coll} = 
\sum_{t=1}^{T} \sum_{i,j \in \mathcal{A}} w_t \cdot \left(1 - \frac{d_t^{i,j}}{d_{penalty}^{i,j}}\right) \cdot M_{i,j}\cdot \mathbb{I}((d_t^{i,j} < d_{penalty}^{i,j} ) \land ( v_i > v_{th}))
\end{equation}
where \(\mathcal{A}\) is the set of all vehicles in the scene, and \(d_t^{i,j}\) and \(d_{penalty}^{i,j}\) represent the distance at time step \(t\) and minimum non‐collision threshold distance while detach the gradient of \(V_j\). \(v_{i}\) is the velocity of the vehicle \(V_{i}\), and 
\(v_{th}\) is a very small velocity threshold. When the vehicle's velocity exceeds \(v_{th}\), it indicates that the vehicle is not in a completely stationary state. This condition ensures that when a moving vehicle is about to collide with a stationary vehicle, the moving vehicle will adjust its trajectory, rather than causing the stationary vehicle to evade the collision, which is a more natural way to prevent collisions. \(M_{i,j}\) is a mask indicating which pairs of agents should evade collisions. In the collision-stage simulation, we configure that all agents, except for the ego and adversarial vehicles, are required to evade collisions. In the evasion-stage simulation, this mask includes all agenets.

To ensure that the vehicles stay within the driving area, we utlilize the following on-road loss,
\begin{equation}
L_{on\_road} = 
\sum_{t=1}^{T} \sum_{p \in P_{offroad}} w_t \cdot \left(1 - \frac{\min_{q \in P_{onroad}} dist(p, q)}{l_{diag}}\right) \cdot  \mathbb{I}(v_i > v_{th})
\end{equation}
where \( P_{offroad} \) and \( P_{onroad} \) are the set of sampled points located off-road and on-road, respectively, within the agent vehicle's bounding box, \( dist(p, q) \) is the Euclidean distance between points \( p \) and \( q \) while detach the gradients of \(q\), and \( l_{diag} \) is the diagonal length of the agent vehicle's bounding box. The gradient of this loss pulls the ego vehicle’s off-road points toward the nearest on-road points, thereby encouraging the denoised trajectory to remain within drivable areas.

\section{Prompt}\label{label detailed}
\begin{figure}[t]
  \centering
  \centerline{\includegraphics[width=\columnwidth]{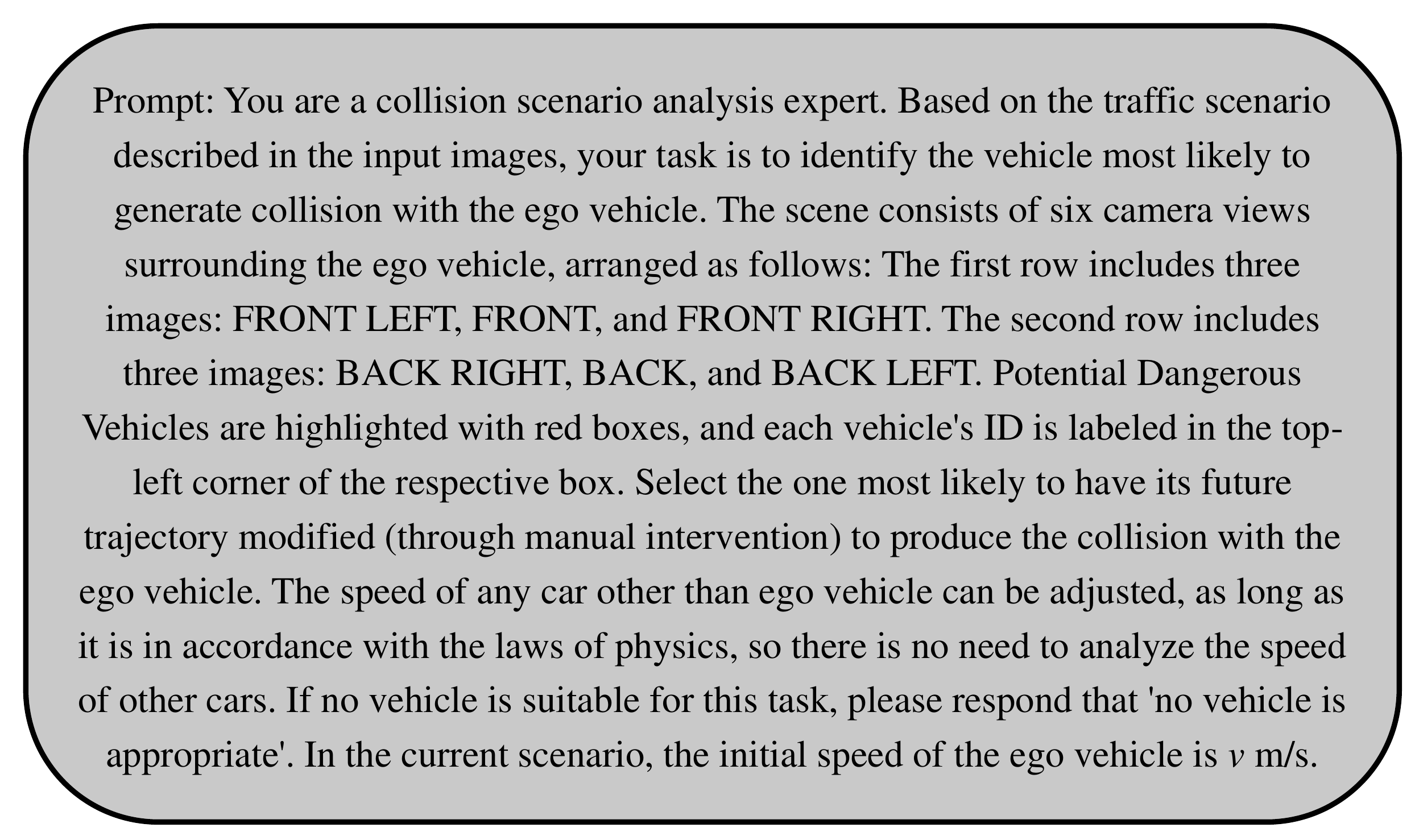}}
  \caption{Original Prompt used in our VLM-based selection.}
  \label{prompt}
\end{figure}

In this section, we present the prompt used in our VLM-based selection of adversarial vehicles in Figure \ref{prompt}.  In the prompt, \(v\) is substituted with the ego vehicle’s velocity in the given scene. For the non-finetuned VLM, we append “Output final answer (number) in <answer> </answer> tags.” at the end to ensure it outputs the vehicle ID for evaluation. For the GRPO-finetuned VLM, we append “Output the thinking process in <think> </think> and final answer (number) in <answer> </answer> tags.” at the end. For the SFT-finetuned VLM, we use the original prompt without any modifications.

\section{Finetuning Setting}\label{VLM-finetuning}
\subsection{VLM finetuning}

\textbf{GRPO-finetuning details}:  We set the learning rate to 0.00002 with a cosine scheduler, enable DeepSpeed Zero3, set the number of generations in GRPO to 6, do not freeze any modules, and follow other settings from the LoRA fine-tuning configuration in \cite{shen2025vlm}. We fine-tune Qwen-VL 2.5 Instruct \cite{qwen2.5-VL} using the GRPO algorithm within the framework of \cite{shen2025vlm} for 2600 steps on 4 A800 GPUs.

\textbf{SFR-finetuing details}: We set the learning rate to 0.00002 with a cosine scheduler, use a gradient accumulation step size of 2, do not freeze any modules, and follow other settings from the LoRA fine-tuning configuration of Qwen-VL 2.5 Instruct in \cite{zheng2024llamafactory}. We fine-tune Qwen-VL 2.5 Instruct \cite{qwen2.5-VL} using the SFT algorithm within the framework of \cite{zheng2024llamafactory} for 2600 steps on a single A800 GPU.

\subsection{Trajectory diffusion model finetuning}
Originally, the context length of the trajectory generation model \cite{zhong2023ctg} is set to 6. Since our framework takes a single-frame initial scene as input, we retrain the model to align with this setup. Following the configuration of \cite{zhong2023ctg}, we introduce two key modifications: (1) the context length is set to 1, and (2) the motion restriction mask for static vehicles is removed to allow more vehicles to collide with the ego vehicle. The trajectory generation model is trained for 80{,}000 steps.

\section{Generated Videos Used for Evaluation.}\label{dataset}
The videos used in our evaluation are generated under a fixed set of 250 samples , randomly selected from the val split, hereafter referred to as the base dataset. The following sections provide the video generated process in each experiment.

\textbf{Videos used in Section \ref{4.2}}: In this section, we compare the generated videos under three baselines: Origin, Naive, and our proposed SafeMVDrive. For the SafeMVDrive set, we apply our full framework to the base dataset and ultimately obtain 41 collision-evasion videos.
For the Origin set, we start from the same 41 initial scenarios used in SafeMVDrive set and convert their original NuScenes trajectories into videos.
For the Naive set, we apply the naive baseline to all 250 initial scenarios in the base dataset and obtain 72 valid collision trajectories, which are then converted into videos.
We evaluate the collision rates of the planner using videos on these three sets.
Since FID scores are empirically affected by the number of images used in evaluation—more images generally lead to lower FID values—for fairness, we randomly sample 41 videos from the Naive set to compute FID.
The videos used in the user study are described in Section~\ref{Human Evaluation}.

\textbf{Videos used in Section \ref{4.4}}: In this section, we compare the generated videos under three methods: Origin,  Collision Stage Only, and Two-Stage Simulation. For the Two-Stage Simulation set, we apply our full framework to the base dataset and ultimately obtain 41 collision-evasion videos.
For the Origin set, we start from the same 41 initial scenarios used in Two-Stage Simulation set and convert their original NuScenes trajectories into videos.
For the Collision Stage Only set, we start from the same 41 initial scenarios used in Two-Stage Simulation set and skip the second simulation to generate videos and eventually get 41 collision videos.
We evaluate the collision rates of the planner using videos and fid on these three sets. The videos used in the user study are described in Section~\ref{Human Evaluation}.


\begin{table}[h]
\centering
\caption{Ablation Study on Loss Functions in the Two-Stage Evasion Trajectory Generator}
\label{tableAB1}
\begin{center}
\begin{small}
\begin{sc}
\resizebox{\columnwidth}{!}{
\begin{tabular}{lcccccc}
\toprule
\textbf{Configuration} & \textbf{CSR} $\uparrow$ & \textbf{ESR} $\uparrow$ & \textbf{Collision Rate} $\downarrow$ & \textbf{Off-road Rate} $\downarrow$ & \textbf{Realism} $\downarrow$ & \textbf{Closest distance} $\downarrow$ \\
\midrule
Whole losses & 0.750 & 0.402 & 0.042 & 0.002 & 0.312 & 5.37\\
$- L_{adv}$ in Collision Stage& 0.471 & 0.703 & 0.034 & 0.000 & 0.308 & 9.11\\
$- L_{no\_coll}$ in Collision Stage& 0.735 & 0.410 & 0.141 & 0.004 & 0.312 & 6.07\\
$- L_{on\_road}$ in Collision Stage& 0.765 & 0.490 & 0.053 & 0.065 & 0.314 & 5.37\\
$- L_{no\_coll}$ in Evasion Stage& 0.770 & 0.127 & 0.024 & 0.000 & 0.313 & 6.32\\
$- L_{on\_road}$ in Evasion Stage& 0.770 & 0.304 & 0.057 & 0.007 & 0.310 & 5.23\\
\bottomrule
\end{tabular}
}
\end{sc}
\end{small}
\end{center}
\end{table}

\section{Ablation Study on Loss Functions in the Two-Stage Evasion Trajectory Generator}\label{loss ablation study}

We conduct ablation studies on the loss functions used in our two-stage evasion trajectory simulator. On 250 validation scenes, we first use the VLM selector to identify safety-critical candidates. Then, we remove one specific loss from the two-stage simulation while keeping the remaining loss terms unchanged to simulate. 

We report the following metrics:

\begin{itemize}
    \item \textbf{Collision Success Rate (CSR)}: the proportion of adversarial vehicles that successfully collide with the ego vehicle during collision simulation. A higher value is better.
    
    \item \textbf{Evasion Success Rate (ESR)}: the proportion of adversarial vehicles that successfully evade during evasion simulation. A higher value is better.
    
    \item \textbf{Collision Rate}: in the final trajectories, the proportion of adversarial vehicles that collide with any vehicle. Since these trajectories are later used for multi-view video simulation and collision cases cannot be rendered, a lower value is preferred. This metric follows the implementation in CTG \cite{zhong2023ctg}.
    
    \item \textbf{Off-Road Rate}: in the final trajectories, the proportion of adversarial vehicles that enter non-drivable areas. A lower value is better. This metric follows the implementation in CTG \cite{zhong2023ctg}.
    
    \item \textbf{Realism}: in the final trajectories, the degree to which the trajectories resemble real-world behavior. In accordance with \cite{zhong2023ctg}, We compare the statistical distribution between simulated trajectories and real-world trajectories. A lower value indicates better realism. This metric follows the implementation in CTG \cite{zhong2023ctg}.
    
    \item \textbf{Closest Distance}: in the final trajectories, the minimum distance between the adversarial vehicle and the ego vehicle, measured by the distance between their center points, which reflects the potential danger level. A lower value is better.

\end{itemize}
The experimental results are shown in Table~\ref{tableAB1}. The results demonstrate that each of our loss terms plays a crucial role. Removing \(L_{adv}\) during the collision stage leads to a higher Closest Distance, indicating a lower safety criticality of the scenes. Removing \(L_{no\_collision}\) results in a higher Collision Rate in the final trajectories. Removing \(L_{on\_road}\) increases the Off-Road Rate in the final trajectories. During the evasion stage, removing \(L_{no\_collision}\) decreases the Evasion Success Rate (ESR), resulting in fewer generated scenarios, while removing \(L_{on\_road}\) similarly increases the Off-Road Rate in the final trajectories. These results verify the rationality and necessity of our loss design.




\section{Hyperparameters study of the losses used in the Two-stage Evasion Trajectory Generator}\label{hyperparemeters of losses}

In this section, we investigate the hyperparameters that control the contributions of different loss terms in the two-stage simulation.The positions of these hyperparameters can be found in Equations (4) and (5) in the main text. In addition to these, we also conduct hyperparameter studies on the weight decay rate factor \(\lambda\). Similar to the previous section, we first use the VLM selector to identify safety-critical candidates on the 250 validation scenes. After that, we vary the hyperparameter corresponding to a specific loss term while keeping the other parameters fixed and then perform the two-stage simulation. We adopt the same evaluation metrics as in the previous section.

The experimental results are shown in Table \ref{table_alpha_collision}, \ref{table_gamma_collision}, \ref{table_beta_collision}, \ref{table_beta_evasion}, \ref{table_gamma_evasion}, and \ref{table_lambda}.
 We vary the hyperparameters controlling the loss contributions with values \{0, 1, 50\}. Overall, setting the value to 0 generally leads to worse performance across various metrics, indicating the necessity of each individual loss term. On the other hand, when the value is within the range of 1 to 50, the differences among the metrics are relatively small, suggesting that our framework is not highly sensitive to hyperparameter selection.

For the weight decay factor \(\lambda\), we evaluate settings of 0, 0.9, and 1. A value of 0 means that the loss is computed using only the prediction at timestamp 1, while a value of 1 averages the loss across all timestamps (i.e., no decay is applied). We observe that the best performance across all metrics is achieved when \(\lambda = 0.9\), which demonstrates the importance of applying a temporal weight decay in our loss design.

\begin{table}[t]
\centering
\caption{Ablation Study on $\alpha$ in Collision Stage}
\label{table_alpha_collision}
\begin{small}
\begin{sc}
\resizebox{\columnwidth}{!}{
\begin{tabular}{lcccccc}
\toprule
\textbf{Configuration} & \textbf{CSR} $\uparrow$ & \textbf{ESR} $\uparrow$ & \textbf{Collision Rate} $\downarrow$ & \textbf{Off-road Rate} $\downarrow$ & \textbf{Realism} $\downarrow$ & \textbf{Closest distance} $\downarrow$ \\
\midrule
$\alpha = 0$ & 0.471 & 0.703 & 0.034 & 0.000 & 0.308 & 9.11 \\
$\alpha = 1$ (default) & 0.750 & 0.402 & 0.042 & 0.002 & 0.312 & 5.37 \\
$\alpha = 50$ & 0.765 & 0.404 & 0.054 & 0.003 & 0.311 & 5.33 \\
\bottomrule
\end{tabular}
}
\end{sc}
\end{small}
\end{table}

\begin{table}[t]
\centering
\caption{Ablation Study on $\beta$ in Collision Stage}
\label{table_beta_collision}
\begin{small}
\begin{sc}
\resizebox{\columnwidth}{!}{
\begin{tabular}{lcccccc}
\toprule
\textbf{Configuration} & \textbf{CSR} $\uparrow$ & \textbf{ESR} $\uparrow$ & \textbf{Collision Rate} $\downarrow$ & \textbf{Off-road Rate} $\downarrow$ & \textbf{Realism} $\downarrow$ & \textbf{Closest distance} $\downarrow$ \\
\midrule
$\beta = 0$ & 0.735 & 0.410 & 0.141 & 0.004 & 0.312 & 6.07 \\
$\beta = 1$ & 0.735 & 0.440 & 0.083 & 0.005 & 0.311 & 5.63 \\
$\beta = 50$ (default) & 0.750 & 0.402 & 0.042 & 0.002 & 0.312 & 5.37 \\
\bottomrule
\end{tabular}
}
\end{sc}
\end{small}
\end{table}

\begin{table}[t]
\centering
\caption{Ablation Study on $\gamma$ in Collision Stage}
\label{table_gamma_collision}
\begin{small}
\begin{sc}
\resizebox{\columnwidth}{!}{
\begin{tabular}{lcccccc}
\toprule
\textbf{Configuration} & \textbf{CSR} $\uparrow$ & \textbf{ESR} $\uparrow$ & \textbf{Collision Rate} $\downarrow$ & \textbf{Off-road Rate} $\downarrow$ & \textbf{Realism} $\downarrow$ & \textbf{Closest distance} $\downarrow$ \\
\midrule
$\gamma = 0$ & 0.765 & 0.490 & 0.053 & 0.065 & 0.314 & 5.37 \\
$\gamma = 1$ (default) & 0.750 & 0.402 & 0.042 & 0.002 & 0.312 & 5.37 \\
$\gamma = 50$ & 0.779 & 0.396 & 0.059 & 0.003 & 0.311 & 5.60 \\
\bottomrule
\end{tabular}
}
\end{sc}
\end{small}
\end{table}

\begin{table}[t]
\centering
\caption{Ablation Study on $\beta$ in Evasion Stage}
\label{table_beta_evasion}
\begin{small}
\begin{sc}
\resizebox{\columnwidth}{!}{
\begin{tabular}{lcccccc}
\toprule
\textbf{Configuration} & \textbf{CSR} $\uparrow$ & \textbf{ESR} $\uparrow$ & \textbf{Collision Rate} $\downarrow$ & \textbf{Off-road Rate} $\downarrow$ & \textbf{Realism} $\downarrow$ & \textbf{Closest distance} $\downarrow$ \\
\midrule
$\beta = 0$ & 0.750 & 0.127 & 0.024 & 0.000 & 0.313 & 6.32 \\
$\beta = 1$ (default) & 0.750 & 0.402 & 0.042 & 0.002 & 0.312 & 5.37 \\
$\beta = 50$ & 0.750 & 0.422 & 0.041 & 0.003 & 0.311 & 4.92 \\
\bottomrule
\end{tabular}
}
\end{sc}
\end{small}
\end{table}

\begin{table}[t]
\centering
\caption{Ablation Study on $\gamma$ in Evasion Stage}
\label{table_gamma_evasion}
\begin{small}
\begin{sc}
\resizebox{\columnwidth}{!}{
\begin{tabular}{lcccccc}
\toprule
\textbf{Configuration} & \textbf{CSR} $\uparrow$ & \textbf{ESR} $\uparrow$ & \textbf{Collision Rate} $\downarrow$ & \textbf{Off-road Rate} $\downarrow$ & \textbf{Realism} $\downarrow$ & \textbf{Closest distance} $\downarrow$ \\
\midrule
$\gamma = 0$ & 0.750 & 0.304 & 0.057 & 0.007 & 0.310 & 5.23 \\
$\gamma = 1$ (default) & 0.750 & 0.402 & 0.042 & 0.002 & 0.312 & 5.37 \\
$\gamma = 50$ & 0.750 & 0.363 & 0.002 & 0.048 & 0.310 & 5.59 \\
\bottomrule
\end{tabular}
}
\end{sc}
\end{small}
\end{table}

\begin{table}[t]
\centering
\caption{Ablation Study on $\lambda$}
\label{table_lambda}
\begin{small}
\begin{sc}
\resizebox{\columnwidth}{!}{
\begin{tabular}{lcccccc}
\toprule
\textbf{Configuration} & \textbf{CSR} $\uparrow$ & \textbf{ESR} $\uparrow$ & \textbf{Collision Rate} $\downarrow$ & \textbf{Off-road Rate} $\downarrow$ & \textbf{Realism} $\downarrow$ & \textbf{Closest distance} $\downarrow$ \\
\midrule
$\lambda = 0$ & 0.640 & 0.011 & 0.182 & 0.091 & 0.336 & 8.61 \\
$\lambda = 0.9$ (default) & 0.750 & 0.402 & 0.042 & 0.002 & 0.312 & 5.37 \\
$\lambda = 1$ & 0.765 & 0.433 & 0.060 & 0.004 & 0.317 & 5.85 \\
\bottomrule
\end{tabular}
}
\end{sc}
\end{small}
\end{table}

\end{document}